\documentclass{article}
\usepackage[preprint]{spconf}
\copyrightnotice{\copyright\ IEEE 2019}
\toappear{To appear in {\it Proc.\ ICASSP 2019,
                    May 12-17, 2019, Brighton, UK.}}
\usepackage{amsmath,graphicx}
\usepackage[hyphens]{url}
\usepackage{subfiles}
\usepackage{color}
\usepackage{booktabs}
\usepackage{amsfonts}
\usepackage{hyperref}
\usepackage{float}
\usepackage{algorithm}
\usepackage[noend]{algpseudocode}
\usepackage{multirow}
\usepackage{array}
\usepackage{subcaption}
\usepackage[dvipsnames]{xcolor}

\title{Phoneme Level Language Models for Sequence Based Low Resource ASR}
\name{Siddharth Dalmia, Xinjian Li, Alan W Black and Florian Metze}
\address{Language Technologies Institute, Carnegie Mellon University; Pittsburgh, PA; U.S.A.\\
\texttt{\{sdalmia|xinjianl|awb|fmetze\}@cs.cmu.edu}}
\begin{document}
\topmargin=0mm
\ninept
\maketitle
\begin{abstract}
Building multilingual and crosslingual models help bring different languages together in a language universal space. It allows models to share parameters and transfer knowledge across languages, enabling faster and better adaptation to a new language. These approaches are particularly useful for low resource languages. In this paper, we propose a phoneme-level language model that can be used multilingually and for crosslingual adaptation to a target language. We show that our model performs almost as well as the monolingual models by using six times fewer parameters, and is capable of better adaptation to languages not seen during training in a low resource scenario. We show that these phoneme-level language models can be used to decode sequence based Connectionist Temporal Classification (CTC) acoustic model outputs to obtain comparable word error rates with Weighted Finite State Transducer (WFST) based decoding in Babel languages. We also show that these phoneme-level language models outperform WFST decoding in various low-resource conditions like adapting to a new language and domain mismatch between training and testing data. 
\end{abstract}
\begin{keywords}
multilingual language models, phoneme-level language models, CTC based decoding, low-resource ASR
\end{keywords}
\section{Introduction}
\label{sec:intro}

There are nearly 7000 languages in the world - some have unique orthography and rules, while some are similar in phonetics, language family, and loan words~\cite{ward1998towards}. With the advent of deep learning, it has been shown that languages can be brought together in a shared universal space with the help of neural networks~\cite{tong2017investigation,vu2014multilingual}. This has multiple advantages like requiring less parameters to model a particular task for many languages, transfer of information across languages in case of low resource languages~\cite{grezl2016study,dalmia2018domain}, cross-lingual adaptation to a new language etc.~\cite{stolcke2006cross,dalmia2018sequence}. However, language models are almost always built to model one language at a time. They are typically word level neural or n-gram language models, which are difficult to share across languages, except when there exist many loan words~\cite{fugen2003efficient, niehues2011wider} or code-switching~\cite{adel2013combination}. 

Collection and cleaning of data in low resource languages can be expensive. Often we find data which is either out-of-domain or is so little that a reliable model cannot be trained using it~\cite{dalmia2018domain}. Building good word language models is also difficult due to various language-specific issues like rich morphology, spelling inconsistencies, etc. If not handled carefully, using these language models to decode an Automatic Speech Recognizer (ASR) usually leads to many out-of-vocabulary words and also bad estimates of uncommon in-domain words. These inconsistencies reduce when observing the data in units smaller than words like characters or phonemes. Hence, there is a need to build language models that can be trained on such smaller units without much preprocessing, and be used with a targeted dictionary during decode time to produce the necessary in-domain words.

In this paper, 
\begin{enumerate}
    \item We propose a phoneme-level language model (Section \ref{sec:prnn}), which is similar to a character-level language model \cite{karpathy2015visualizing,maas2015lexicon} in terms of the granularity of training unit. Additionally, this can be used to create a shared representation using the language agnostic units, International Phonetic Alphabet. It allows us to share language model parameters across multiple languages. We show that we can build multilingual phoneme-level language models where we get the same perplexity on all languages without increasing the number of parameters, effectively using six times fewer trainable parameters. We see considerable reduction in perplexity during crosslingual adaptation of our multilingual language model versus a monolingual language model (Section \ref{sec:crosslingual}). 

    \item These phoneme-level language models can be used to decode CTC acoustic model outputs by doing a prefix tree based beam search, a slight modification to open vocabulary beam search~\cite{maas2015lexicon} and prefix tree based search~\cite{hannun2014first}. We show that on an average they perform around 6.1\% better than using open vocabulary character-based decoding~\cite{maas2015lexicon,zenkel2017comparison} and are at par with the popularly used WFST-based decoding~\cite{miao2015eesen}(Section \ref{sec:ctcdecode}). 

    \item We find that our approach is better in low resource scenarios and decoding with monolingual and (adapted) multilingual phoneme-level language models both performing better than WFST decoding (Section \ref{sec:crosslingual}). We also show that these models are robust towards domain mismatch and can be trained with only out-of-domain data, Bible text, and be decoded by just providing a list of in-domain words, conversational speech.
\end{enumerate}

We start by explaining the related work and datasets which we used in this paper. Section \ref{sec:prnn} explains our proposed approach, followed by experiments and results in Section \ref{sec:experiments} showing the aforementioned contributions.

\section{Related Work and BABEL Dataset}

Building character-level n-gram language models is difficult due to the need for long contexts. However, with the use of RNNs and (Long Short Term Memory) LSTMs, it has been shown that character-level RNN language models (CLMs) can produce sentences that are semantically and syntactically correct \cite{karpathy2015visualizing} due to their capability of capturing long contexts \cite{sutskever2011generating}. While most tasks in natural language processing that use CLMs build word-level language models \cite{bojanowski2016enriching,tsvetkov2016polyglot}, in speech recognition, especially for decoding sequence-based acoustic models, we are interested in making character-level sentence language models, where \texttt{space} is considered as another character~\cite{karpathy2015visualizing,hwang2016character,maas2015lexicon}. 

Parameter sharing in language models has been done in the past with words as units for code-switching~\cite{adel2013combination}. Though these models tend to benefit by having common loan words~\cite{fugen2003efficient, niehues2011wider}, there are very few common target units which makes it harder to share parameters. A polyglot language model proposed in \cite{tsvetkov2016polyglot} tries to approach some of the problems mentioned above. However, this work was limited to producing only words and did not explore its capabilities at sentence level (the model cannot produce sentences). Though this work made interesting observations in deciding what languages to use for improving performance of multilingual models, they didn't explore its capability to adaptation towards a new target language.

There are multiple ways to decode CTC acoustic models. One of the simplest ones is greedy decoding where we choose the top prediction at every frame and apply the CTC squash function to get the target sequence, as shown in ~\cite{graves2006connectionist}. We can get better predictions by just doing a prefix based beam search without any language models where we add paths to the beam that lead to a valid word, ~\cite{hannun2014first,graves2014towards}. CTC being a conditionally independent model, benefits a lot when decoded with a language model, this was shown using both character \cite{maas2015lexicon,hwang2016character} and word \cite{hannun2014first,miao2015eesen} language model. A word language model can be applied at every word boundary during a prefix based search \cite{hannun2014first} or by composing a \texttt{TLG} graph using WFST \cite{miao2015eesen} and a character language model can be used while inserting a character \cite{maas2015lexicon,hwang2016character}. While the first one produces words from a fixed lexicon, it requires us to make word-level language models. On the contrary, the latter requires easier to train character language models but performs open-vocabulary decoding which may not always be optimal, especially in low resource scenarios. 

In this paper, we use 9 different languages from the IARPA BABEL Research Project (IARPA-BAA-11-02). We choose three languages, Cebuano, Mongolian and Amharic, as our unseen test languages and choose two other languages spoken in nearby regions from each of these three, i.e., Javanese, Tagalog, Turkish, Kazakh, Swahili, and Zulu, as our training languages. We hope to maximize closeness in language family and loan words by using this heuristic. Table \ref{tab:dataset} summarizes the number of phonemes, amount of training data and the out-of-vocabulary (OOV) rate for the languages we used in our experiments on the “Full Language Pack” (FLP) condition~\footnote{This work used releases IARPA-babel105b-v0.4, IARPA-babel106-v0.2g, IARPA-babel202b-v1.0d, IARPAbabel204b-v1.1b, IARPA-babel206b-v0.1d, IARPAbabel301b-v2.0b, IARPAbabel302b-v1.0a, IARPAbabel307b-v1.0b, IARPAbabel401b-v2.0b and IARPAbabel402b-v1.0b provided by IARPA BABEL Research Program}.

\begin{table}[H]
\centering
\begin{tabular}{ccccc}
\toprule
                                          & Languages & \#Units & \#Train Utts &  OOV\% \\
\midrule
\multirow{3}{*}{Test}                     & Cebuano   & 28      & 42k       & 3.7       \\
                                          & Mongolian & 50      & 45k       & 4.5       \\
                                          & Amharic   & 58      & 41k       & 11.3       \\
\midrule
\multirow{6}{*}{Train}                    & Javanese  & 31      & 46k       & 4.4       \\
                                          & Tagalog   & 25      & 93k       & 2.8       \\
                                          & Turkish   & 29      & 81k       & 5.7       \\
                                          & Kazakh    & 39      & 48k       & 6.1       \\
                                          & Swahili   & 37      & 44k       & 7.7       \\
                                          & Zulu      & 44      & 60k       & 13.4      \\
\bottomrule                                          
\end{tabular}
\caption{Overview of the FLP Babel Corpora used in this work.}
\label{tab:dataset}
\end{table}

\section{Phoneme Level Language Models}
\label{sec:prnn}

\textbf{Character Language Models (CLMs):} A typical CLM involves an embedding lookup, \texttt{Emb} $\in \mathbb{R}^{d \times |V|}$, for each character $c \in \{V\}$ to an embedding dimension $d$, \texttt{LSTMs} for modeling past context and a final output transformation $\mathbf{W}_{out}$ which is passed to a \texttt{softmax} to produce distribution over $V$. 
Given a sequence $\mathbf{c}_{1 \ldots t-1}$, the distribution over the next character of the sequence is then computed as follows:
 \begin{align*}
 p(c_t \mid c_{1 \ldots t-1}) = \texttt{softmax}(\mathbf{W}_{out}\texttt{LSTM}(
 \texttt{Emb}(c_{1 \ldots t-1})) + \mathbf{b}_{out})
 \end{align*}
\noindent\textbf{Phoneme Language Models (PLMs):} In this paper, we are interested in building such CLMs but on language universal characters, also called as the International Phonetic Alphabet~(IPA). We can convert the words of any language into their corresponding IPA symbol, by using any rule-based grapheme-to-phoneme (G2P) library. In this paper, we use the Epitran \cite{mortensen2018epitran} G2P library. Although using phonemes comes with the added cost of a G2P library, typically speech recognizers built using phonemes have better performance when compared to character-level models~\cite{zenkel2017comparison}. 

\noindent\textbf{Multilingual Phoneme Language Models (Multi-PLMs):} Another key advantage of using phonemes is that with IPA, we enter a language agnostic space, allowing us to use a single model to decode sequences in multiple languages. Authors in \cite{tsvetkov2016polyglot} show that incorporating a language tag while training the RNN language model helps to improve multilingual language models. Since we want to use our model to work in crosslingual adaptation scenarios, we want to bring all the phonemes in same space. We modify the model proposed by \cite{tsvetkov2016polyglot} to provide language identification only for sentence and word boundaries. Effectively, the input units ($x$) are sum of the union of all the phonemes $\phi$ in each of the languages~($\phi_{l}$) and language specific \texttt{<space>} and \texttt{<sos>}. 

Further, we also introduce a ``masked-training'' approach to improve the model training by computing softmax and calculating loss only on units belonging to the languages being trained. The basic motivation behind this approach is to bring the advantages of the ``block-softmax'' approach that has been very useful for multilingual acoustic models ~\cite{scanzio2008use,vesely2012language,heigold2013multilingual,dalmia2018sequence} into a ``shared-softmax'' model. 
\begin{align*}
&\texttt{lang\_mask}_l = \begin{cases} 
      \texttt{True}  & \text{if } \mathbf{x} \in \{\phi_{l}\} \\
      \texttt{False}  & \text{if } \mathbf{x} \notin \{\phi_{l}\} 
\end{cases}\\
&\texttt{ind} = \texttt{where}(\texttt{lang\_mask} = \texttt{True})\\
&\texttt{logits} = \mathbf{W}_{out}\texttt{LSTM}(\texttt{Emb}(x_{1},\ldots,x_{t-1})) + \mathbf{b}_{out}\\
&\texttt{sparse\_softmax} = \texttt{softmax}(\texttt{gather}_{\texttt{ind}}(logits))
\end{align*}
This approach also ensures that only language-specific gradient flows through the network for any training example, thereby not penalizing the model on distributing activations on invalid phones for any training example. We found that the ``masked-training'' approach not only gives better results but also helps in faster convergence. 

\section{EXPERIMENTS AND OBSERVATIONS}
\label{sec:experiments}
\subsection{Multilingual Phoneme Language Model}
We build PLMs on each of the training languages and compare their performance with a Multi-PLM built on all the training languages put together. The multilingual model uses the same number of parameters as the models built for individual languages, effectively using six times fewer parameters while fulfilling the same purpose for each of the six languages. Both the models use a single layer LSTM with 1024 hidden units and 64 dimensional embeddings. All models have a dropout of $0.4$ in the LSTM layers and are implemented using Tensorflow. These models are chosen after a parameters search on different embedding sizes (64, 128, 256), hidden units (256, 512, 1024) and dropout rate (0.4, 0.2, 0).

\subsubsection{Parameter Reduction using Multi-PLMs}

Table \ref{tab:lm} shows the phoneme-level perplexity results (not counting sentence boundaries) of the multilingual model and compares it with corresponding monolingual models. It shows that the Multi-PLM model matches the performance of the PLM Large model while using roughly $6$ times lesser parameters. For comparison, we also show the perplexities obtained using a smaller PLM model which uses roughly $1/6$ effective trainable parameters (obtained by using LSTM with 256 units instead of 1024 and a dropout of 0 instead of 0.4). We can see that the Multi-PLM does better than the PLM Small model and showing that the model is benefiting from learning a shared representation.

\begin{table}[H]
\centering
\begin{tabular}{cccc}
\toprule
            & PLM       & PLM     & Multi-PLM \\
            & Small     & Large   & Large \\
\midrule
\# Params   & $\sim$0.4M$\times$6  & $\sim$4.5M$\times$6    & \textbf{$\sim$4.6M} \\
\midrule
Javanese    & 3.91      & 3.80 & 3.80   \\
Tagalog     & 3.62      & 3.43 & 3.46   \\
Turkish     & 3.53      & 3.36 & 3.38   \\
Kazakh      & 3.02      & 2.89 & 2.89   \\
Swahili     & 3.63      & 3.44 & 3.50   \\
Zulu        & 4.18      & 3.95 & 4.00   \\
\bottomrule
\end{tabular}
\caption{PLM (Small and Large) and Multi-PLM (Large) perplexities for different languages in the training set.}
\label{tab:lm}
\end{table}

\subsubsection{Crosslingual Adaptation using Multi-PLMs}

\begin{figure}[H]
  \centering
  \includegraphics[width=0.95\linewidth]{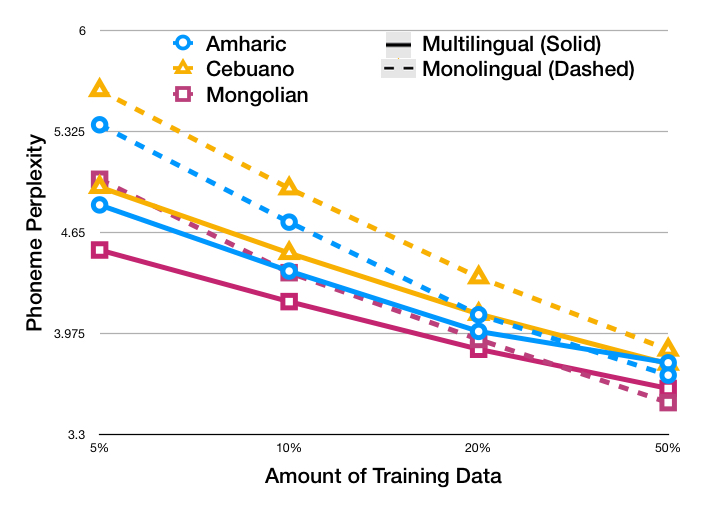}
  \caption{PPL after adaptation of Multi-PLM to target languages on different amounts of data. Multi-PLM outperforms PLM for small amounts of training data.}
  \label{fig:crosslm}
\end{figure}

Here, we compare the performance of our monolingual and multilingual PLMs by adapting them to various amounts of training data in a target language. We use Amharic, Cebuano and Mongolian as our test crosslingual languages; the multilingual model has not seen any of these languages before adaptation. We train the monolingual model from scratch without using any adaptation. Figure \ref{fig:crosslm} shows that the Multi-PLM adapts better to the target language when compared to a PLM trained in that language. The gains are consistent across languages, and they reduce as the amount of training data increases in the target language. When the model has seen 50\% of the training data, the gains seem to disappear, and for some languages, the PLM performs better than the adapted multilingual model.

\subsection{Decoding using Phoneme LM}
\label{sec:ctcdecode}
Previous works that employ character-level language models are typically deployed in the context of high resource scenarios~\cite{zenkel2017comparison,maas2015lexicon,hwang2016character}, for example, \cite{zenkel2017comparison} trains on around 112M characters. However, it is often impossible to collect and clean such large amounts of data in low-resource languages, and these open vocabulary decoding approaches do not perform well. We think this is because the language models that are trained might not reliably output valid OOV words. For example, from the experiments shown in Table~\ref{tab:plm} for Zulu, open vocabulary CLM decoding outputs $4k$ incorrect OOV words out of the $46k$ total words in the hypothesis. 

To solve this issue, we combine the CLM based beam search decoding with a prefix tree to restrict the paths taken in the lattice during beam search to only valid words, similar to \cite{hannun2014first,graves2014towards}. In Table \ref{tab:plm}, we show the results of in-vocabulary decoding of PLMs and compare them with the previous works, i.e., open vocabulary decoding using CLMs~\cite{zenkel2017comparison,maas2015lexicon} and WFST based decoding~\cite{mortensen2018epitran,miao2015eesen} with CTC based acoustic models. We see that PLM based decoding does much better than open vocabulary CLM decoding. These improvements stem either from the ability to build higher quality acoustic models using phonemes instead of characters, or in-vocabulary decoding, both of which our proposed model facilitates over prior work. We also observe that this approach performs comparatively with WFST based decoding.
\begin{table}[H]
\centering
\begin{tabular}{c|ccc}
\toprule
Babel     & WFST & CLM & PLM \\
Languages & \multicolumn{3}{c}{Based Decoding}\\
\midrule   
Cebuano   & 57.1 & 71.1  & 67.9          \\
Mongolian & 60.5 & 84.3  & \textbf{59.0}    \\
Amharic   & 57.2 & 64.8  & \textbf{57.6}  \\
\midrule   
Javanese  & 65.7 & 68.4  & \textbf{64.8} \\
Tagalog   & 55.7 & 58.0  & \textbf{55.8} \\
Kazakh    & 57.8 & 64.2  & \textbf{61.3}         \\
Turkish   & 56.9 & \textbf{58.5}  & 59.4          \\
Swahili   & 61.2 & \textbf{50.7}  & \textbf{50.8} \\
Zulu      & 65.2 & 75.3 & \textbf{63.7} \\
\bottomrule
\end{tabular}
\caption{\% WER for each of the languages used using different kinds of decoding strategies; WFST decoding using word language models, open-vocabulary decoding using CLMs and in-vocabulary decoding using PLMs. Almost always, PLM based decoding performs better than CLMs and as good as WFST.}
\label{tab:plm}
\end{table}

We only use the training lexicon as the in-vocabulary dictionary while decoding PLMs. For training the acoustic models for CLM and PLM decoding, we add an extra target token between every pair of words, representing a word boundary. During our initial experiments, we found that language model weight of $1.0$, insertion penalty of $0.35$ and beam size of $40$ worked best for our development set and we used this value for all our experiments. For WFST, we use a beam size of $9.0$, lattice beam of $4.0$ and acoustic model weight of $0.6$. Note that the numbers shown for WFST are the best WER found using various word insertion penalties, n-gram language model parameters and prior decoding for each language. Whereas for the PLM based model, we use fixed parameters across languages, presenting room for further improvements.

\subsubsection{Comparing Decoding Strategies for Crosslingual Adaptation}
\label{sec:crosslingual}

We now compare the decoding results of using PLMs and (adapted) Multi-PLMs against a word-based WFST decoding. Again, we use Amharic, Cebuano and Mongolian as our test crosslingual languages. For this experiment, we train a monolingual CTC acoustic model, which is a 2 layer bidirectional LSTM with 360 units, and use it with the PLMs and WFSTs to decode the CTC acoustic models. Table \ref{my-label} shows the word error rates for various amounts of training data on different acoustic models. We can see that our monolingual as well as multilingual language model decoding does better than WFST based decoding most of the times. We also notice that for low resource scenarios, getting a good prior estimate is not possible and hence the WFST results fluctuate depending on the language and the data sub-selection. 

\begin{table}[H]
\centering
\begin{tabular}{c|c|cccc}
\toprule
Crosslingual & Decoding & \multicolumn{4}{c}{Training Data (\%)} \\
Languages                  & Strategies & 5\%            & 10\%           & 20\%           & 50\% \\
\midrule
\multirow{3}{*}{Amharic}   & WFST & 89.94          & 87.93          & 82.84          & 78.02          \\
                           & PLM & \textbf{86.50} & \textbf{82.66} &   \textbf{78.18}   & \textbf{69.90}     \\
                           & Multi-PLM & \textbf{86.29} & \textbf{82.07} & \textbf{78.30} & \textbf{70.91} \\
\midrule
\multirow{3}{*}{Cebuano}   & WFST & 92.96          & 91.29          & 88.36          & 83.69          \\
                           & PLM &    \textbf{89.85}            & \textbf{86.70} & \textbf{82.72}          & \textbf{77.23}   \\
                           & Multi-PLM & \textbf{89.85} & \textbf{86.04} & \textbf{82.53} & \textbf{77.23} \\
\midrule
\multirow{3}{*}{Mongolian} & WFST & 91.07          & \textbf{83.83} & 88.42          & 80.61    \\
                           & PLM & \textbf{88.12} & \textbf{84.89} & \textbf{81.00} & \textbf{73.19} \\
                           & Multi-PLM & \textbf{88.05} & \textbf{84.98} & \textbf{80.96} & \textbf{73.33} \\
\bottomrule
\end{tabular}
\caption{Comparing decoding strategies on crosslingual adaptation with different amounts of training data. We see that PLM and Multi-PLM based decoding does better than WFST in almost all cases.}
\label{my-label}
\end{table}

\begin{table}[H]
\centering
\begin{tabular}{c|cccc}
\toprule
Probability of & \multicolumn{4}{c}{Training Data (\%)} \\
Improvement & 5\% & 10\% & 20\% & 50\% \\
\midrule
Amharic & \textbf{97.9} & \textbf{100.0} & 10.3 & 0.0 \\
Cebuano & \textbf{99.9} & \textbf{100.0} & \textbf{97.5} & \textbf{95.3} \\
Mongolian & \textbf{83.7} & 11.8 & \textbf{67.6} & 5.7\\
\midrule
\end{tabular}
\caption{Bootstrap comparison of Multi-PLM with PLM based decoding to estimate the prob. of improvement at 95\% conf. interval.}
\label{tab:sig}
\end{table}

Though language model perplexity improvements are seen using the crosslingually adapted multilingual language models, the improvements are not apparent in terms of word error rates. To understand the ``significance of improvement'' made using the Multi-PLM decoding over the PLM decoding, we use the bootstrapped tests presented in \cite{bisani2004bootstrap} using the Kaldi \texttt{compute-wer-bootci} tool. It returns a probability estimate of improving WER of system $1$, here a Multi-PLM, by bootstrapping it with system $2$, a monolingual PLM. Table \ref{tab:sig} shows that the improvements are significant, always for 5\% of the data and almost always for the rest.

\subsubsection{Domain Robustness of Decoding Strategies}
\label{sec:domain}

We finally compare the robustness of the two best decoding strategies on domain mismatched conditions. Here, we assume the Bible as one of the sources of text in any low resource language and use it to train our language model. For generating the lexicon for both PLMs and WFSTs, we run the words of the Bible through Epitran G2P library. For the WFST, we train multiple n-gram language models with various discounting and choose the one that performs the best for our in-domain validation data. We then use this language model along with an in-domain acoustic model and decode it using the two strategies using the in-domain target dictionary. From Table \ref{tab:domain} we can see that PLM based decoding performs much better than WFST based decoding showing its capability of generating words outside language model training data by just using a targeted lexicon.

\begin{table}[H]
\centering
\begin{tabular}{c|cc}
\toprule
Babel     & WFST & PLM \\
Languages & \multicolumn{2}{c}{Based Decoding}\\
\midrule   
Cebuano   & 86.2  & \textbf{79.8}  \\
Javanese  & 93.1  & \textbf{80.8}  \\
Tagalog   & 83.4  & \textbf{68.9}  \\
Kazakh    & 78.3  & \textbf{72.5}  \\
\bottomrule
\end{tabular}
\caption{\% WER for languages using different decoding strategies on LMs trained on the Bible text. We see that in-vocabulary decoding using PLMs does much better than WFST based decoding.}
\label{tab:domain}
\end{table}

\section{Conclusion}
\label{sec:conclusion}

In this paper, we propose a phoneme-level language model and present a unique way of training it multilingually thereby using around six times fewer parameters without much increase in perplexity. We show that it is beneficial to use multilingual models when adapting to a new language in very low resource settings. As the amount of training data increases, the monolingual PLM starts to outperform the multilingually adapted models. 

We show a way of using phoneme-level language models to decode CTC acoustic models using a targetted lexicon which gives us significant gains over open-vocabulary decoding using CLMs and comparable results to the traditional WFST based decoding. We show that our approach outperforms WFST in low resource conditions, like crosslingual adaptation, where building good n-gram word language models is hard.

Finally, we explore the domain adaptation capabilities of our model, where we train on words that are outside the target domain. We exploit the phoneme-level training of the language model coupled with our targeted lexicon decoding approach to improve the robustness of our model.
\section{ACKNOWLEDGEMENTS}
\label{sec:ack}
This project was sponsored by the Defense Advanced Research Projects Agency (DARPA) Information Innovation Office (I2O), program: Low Resource Languages for Emergent Incidents (LORE-LEI), issued by DARPA/I2O under Contract No. HR0011-15-C-0114. This work used the Extreme Science and Engineering Discovery Environment (XSEDE), which is supported by National Science Foundation grant number OCI-1053575. Specifically, it used the Bridges system, which is supported by NSF award number ACI-1445606, at the Pittsburgh Supercomputing Center (PSC).

\bibliographystyle{IEEEbib}
\bibliography{refs}

\end{document}